%% file: main.tex
\documentclass[sigconf, 9pt]{acmart}

\newif\ifedit
\edittrue

\ifedit
    \newcommand{\tp}[1]{
        \vskip 1ex\noindent
        \colorbox{red}{
            \parbox{\columnwidth - 2\fboxsep}{
                \textbf{Issues:} #1
            }
        }
    }
    \newcommand{\preq}[1]{
        \vskip 1ex\noindent
        \colorbox{red}{
            \parbox{\columnwidth - 2\fboxsep}{
                \textbf{Pre-requisites:} #1
            }
        }
    }
    \newcommand{\kaiyuan}[1]{\textcolor{red}{KY: #1}}
    \newcommand{\scott}[1]{\textcolor{red}{MZ: #1}}
    \newcommand{\stephen}[1]{\textcolor{red}{SX: #1}}

    \definecolor{updatecolor}{RGB}{0,102,204} 
    \definecolor{deletecolor}{RGB}{255,0,0}   
    \definecolor{addcolor}{RGB}{0,153,0}      
    
    \newcommand{\delete}[1]{{\color{deletecolor}\sout{#1}}}
    \newcommand{\add}[1]{{}#1}
    \newcommand{\todo}[1]{\ClassWarning{NOT READY TO SUBMIT}{There is something left todo} \textcolor{red}{[TODO: #1]}}

\else
    \newcommand{\tp}[1]{}    
    \newcommand{\preq}[1]{}    
    \newcommand{\kaiyuan}[1]{}
    \newcommand{\scott}[1]{}
    \newcommand{\stephen}[1]{}

    \newcommand{\delete}[1]{}
    \newcommand{\add}[1]{#1}

    \newcommand{\todo}[1]{\ClassWarning{NOT READY TO SUBMIT}{There is something left todo}}

\fi

\setcopyright{acmlicensed}\acmConference[ACM Sensys '25]{The 23rd ACM Conference on Embedded Networked Sensor Systems}{May 6--9, 2025}{Irvine, California, USA}
\copyrightyear{2025}
\acmYear{2025}
\acmBooktitle{The 23rd ACM Conference on Embedded Networked Sensor Systems (ACM Sensys '25), May 6--9, 2025, Irvine, California, USA}
\acmDOI{}
\acmISBN{}
\usepackage{url}
\usepackage{xcolor}
\usepackage{color}
\usepackage{graphics}
\usepackage{graphicx}
\usepackage{multirow}
\usepackage{xspace}
\usepackage{enumitem}
\usepackage{subcaption}
\usepackage{balance}
\usepackage{acmart-taps}
\usepackage{booktabs}

\usepackage[subtle]{savetrees}
\begin{document}

\title[SHADE-AD: An LLM-Based Framework 
for Synthesizing Activity Data of Alzheimer’s Patients]
{
SHADE-AD: An LLM-Based Framework \\
for Synthesizing Activity Data of Alzheimer’s Patients}
\author{Heming Fu$^{1, 2}$, Hongkai Chen$^2$, Shan Lin$^1$\textsuperscript{\dag}, Guoliang Xing$^2$\textsuperscript{\dag}}
\affiliation{%
 \institution{$^1$Stony Brook University, $^2$The Chinese University of Hong Kong}
 \country{\{heming.fu, shan.x.lin\}@stonybrook.edu, \{hkchen, glxing\}@ie.cuhk.edu.hk}
}
\authornote{\textsuperscript{\dag}Co-Corresponding Authors.}

\renewcommand{\shortauthors}{H. Fu et al.}


\input{sections/abstract}

\maketitle

\input{sections/intro}
\input{sections/related}
\input{sections/system}
\input{sections/validation}

\input{sections/result}
\input{sections/conclusion}
\input{sections/acknowledgment}



\balance
\bibliographystyle{ACM-Reference-Format}
\input{main.bbl}


\end{document}

%% file: sections/abstract.tex
\begin{abstract}

Alzheimer’s Disease (AD) has become an increasingly critical global health concern, which necessitates effective monitoring solutions in smart health applications. However, the development of such solutions is significantly hindered by the scarcity of AD-specific activity datasets.  
To address this challenge, we propose SHADE-AD, a Large Language Model (LLM) framework for \textbf{S}ynthesizing \textbf{H}uman \textbf{A}ctivity \textbf{D}atasets \textbf{E}mbedded with \textbf{AD} features.
Leveraging both public datasets and our own collected data from 99 AD patients, SHADE-AD synthesizes human activity videos that specifically represent AD-related behaviors. By employing a three-stage training mechanism, it broadens the range of activities beyond those collected from limited deployment settings. 
We conducted comprehensive evaluations of the generated dataset, demonstrating significant improvements in downstream tasks such as Human Activity Recognition (HAR) detection, with enhancements of up to 79.69\%. Detailed motion metrics between real and synthetic data show strong alignment, validating the realism and utility of the synthesized dataset. These results underscore SHADE-AD’s potential to advance smart health applications by providing a cost-effective, privacy-preserving solution for AD monitoring.



\end{abstract}

%
%



%% file: sections/intro.tex
\section{Introduction} \label{sec:intro}

Alzheimer’s Disease (AD) is a rapidly escalating global health crisis, affecting approximately 6.9 million Americans age 65 and older in 2024, representing about 10.9\% of that age group~\cite{alzheimers2024facts}. This number is projected to grow to 13.8 million by 2060, highlighting an urgent need for effective monitoring and early intervention strategies~\cite{Kourtis2019}. 

Smart health technologies, powered by artificial intelligence (AI) and sensor systems, offer promising solutions by enabling continuous, non-invasive monitoring of patients’ daily activities~\cite{ouyang2024admarker}. These technologies are crucial for facilitating early diagnosis and personalized care, which can significantly improve patient outcomes and reduce healthcare costs~\cite{prince2014world, emomarker, drhouse, artfl}. However, the effectiveness of AI models in these applications depends heavily on the availability of high-quality, domain-specific datasets for training~\cite{esteva2019guide, dataset}. Without such datasets, AI models may fail to accurately recognize AD-specific behaviors, leading to delayed diagnoses, inadequate patient monitoring, and potential safety risks~\cite{senders2018introduction, WenliZhang}.

Despite the prevalence of the Alzheimer's Disease, AD-specific behavioral datasets are very limited. Existing datasets often lack the necessary domain-specific features to capture the nuanced behaviors associated with AD patients, such as subtle changes in gait, posture, and movement patterns indicative of cognitive decline~\cite{brainsci9020034}. Using generative models to augment the AD behavior dataset presents a potential solution to address the aforementioned limitation~\cite{electronics13183671,inproceedings, adclip}. However, though existing synthetic data generation models, such as those leveraging vision-language models like BERT~\cite{bertasius2021space}, Stable Diffusion~\cite{rombach2022high}, and MotionDiffuse ~\cite{zhang2022motiondiffuse} have demonstrated strong capabilities in understanding and generating general visual content~\cite{LLMunderstanding,li2023synthetic}, they are primarily trained on data from healthy individuals and do not capture the specific motor and behavioral anomalies seen in AD patients~\cite{liu2024datasetslargelanguagemodels}. For example, these models may generate generic walking patterns but fail to reproduce the shuffling gait or balance issues common in AD. This limitation hinders the development of effective AI models for AD monitoring, as models trained on general datasets may not accurately detect or interpret AD-specific behaviors. 

A further challenge lies in validating disease-specific synthetic data, particularly due to the complexity of patient-specific physical characteristics. In the case of AD, there is currently no established model or standardized metrics capable of accurately capturing the wide range of behaviors associated with cognitive and motor decline~\cite{FanYi2023, Kourtis2019}. Although several studies on gait analysis provide valuable insights~\cite{Zhaoying2023}, they predominantly focus on lower-body movements and lack a comprehensive approach to the subtle, full-body action patterns that characterize AD.

\begin{figure}[htbp]
  \centering
  \vspace{-0.5em}
  \includegraphics[width=\linewidth]{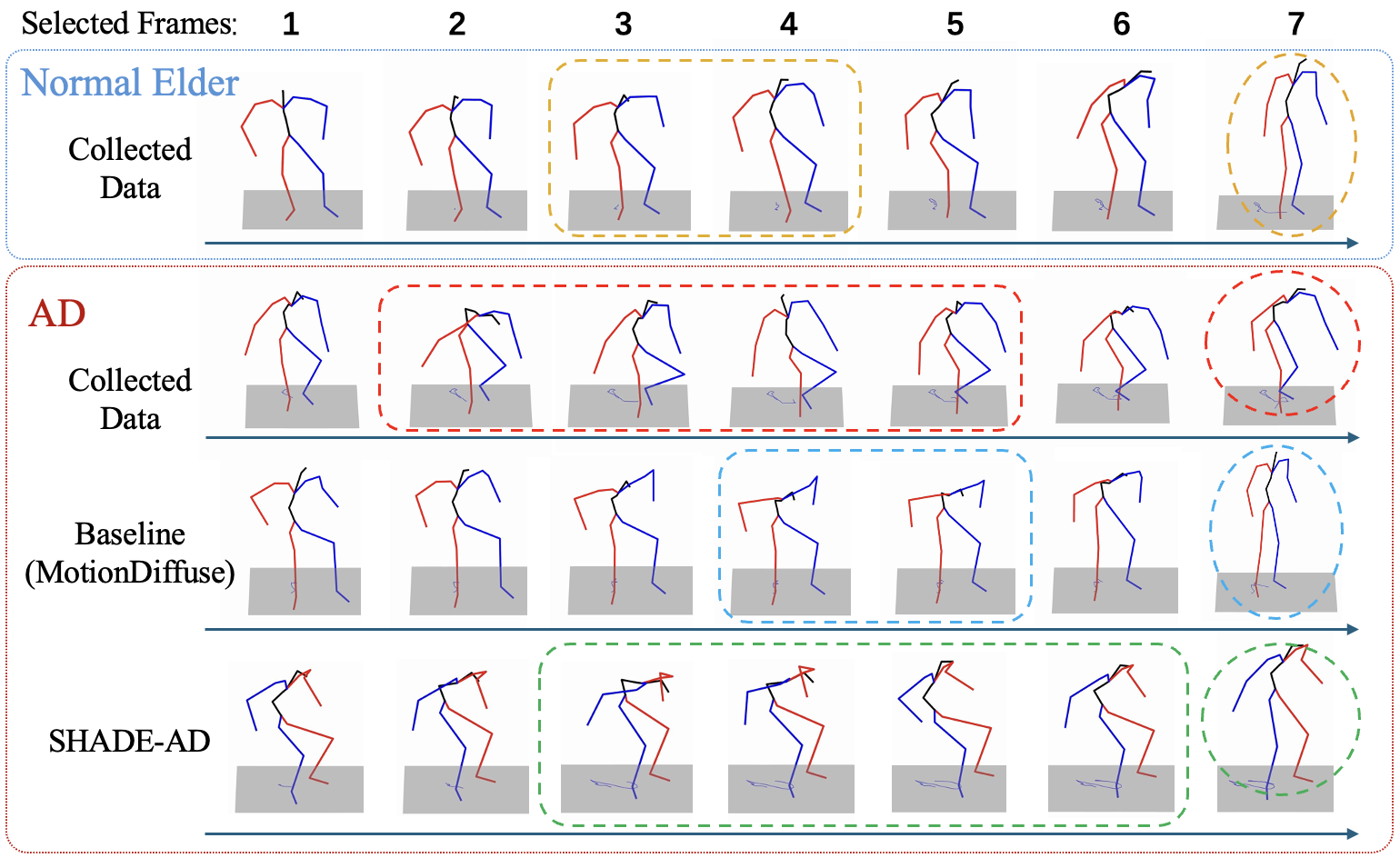}
  \captionsetup{skip=0pt}
  \caption{Comparison of ``standing up from a chair'' motion highlighting two key differences: (1) Standing transition (shown in rectangles): AD patients exhibit oscillation during the standing process (red), while Normal Elder (yellow) and Baseline (blue) both miss this characteristic and demonstrate smooth standing movement. SHADE-AD successfully reproduces this AD-characteristic oscillation in frames 3-6 (green). (2) Final posture (shown in circles): AD patients maintain a forward stooped posture after standing (red), whereas Normal Elder (yellow) and Baseline (blue) both miss this feature and achieve an upright stance. SHADE-AD accurately captures this forward stoop characteristic of AD patients (green).}
  \label{fig:example}
  \vspace{-1.2em}
\end{figure}

Figure~\ref{fig:example} shows that AD patients exhibit oscillatory movements when standing up and maintain a stooped posture afterward, unlike normal elders and other motion generative models, such as MotionDiffuse~\cite{zhang2022motiondiffuse} (prompted with “an Alzheimer’s Disease patient standing up from a chair”), which generate a smooth transition to an upright position, much like a healthy subject.

To address these challenges, we develop SHADE-AD, a generative model to synthesize activity videos of AD patients.
SHADE-AD embeds AD features through a three-stage process. The first two stages develop feature extraction abilities for AD terminology and action patterns. In the third stage, SHADE-AD incorporates motion metrics derived from 12 key joints, which validate the realism and clinical relevance of the synthetic dataset.

In summary, the key contributions of our work are:
\begin{itemize}
\item We propose a generative model, SHADE-AD, for synthetic 3D skeleton activity videos specific to AD patients, addressing the scarcity of AD behavior data.
\item We design a framework that embeds AD-specific behavioral knowledge into a CLIP-based LLM encoder, enhancing the model’s ability to generate the nuanced actions of AD patients.
\item We validate the effectiveness of the synthetic dataset generated by SHADE-AD using motion metrics of 12 key joints, and demonstrate its ability to significantly improve performance across multiple downstream tasks in smart health.
\end{itemize}

%% file: sections/related.tex
\section{Related Work}\label{sec: related}

The development of effective AI models for smart health applications, particularly in AD scenarios, relies heavily on the availability of comprehensive behavioral datasets. This section reviews the existing literature on AD behavior datasets, data augmentation methods, and the use of LLMs for data generation, highlighting the gaps that this work aims to address. We will compare these methodologies with our approach in the preliminary results part.

\vspace{-0.5em}
\subsection{Behavioral Datasets}

Behavioral datasets are crucial in disease detection and patient monitoring applications, as they provide insights into patient activities that can indicate the onset or progression of various conditions~\cite{ouyang2024admarker}. 
However, there is a significant lack of behavioral datasets that encompass a wide range of scenarios and activities for disease specific patient groups.
Most existing datasets focus on clinical data such as neuroimaging, genetic markers, and cognitive assessments~\cite{markerreivew}, while neglecting real-world behavioral data. Some initiatives have attempted to collect patient-related behavior through sensor data and video recordings~\cite{harmony}, but these are limited in scope and often do not cover the diverse activities encountered in daily living. The scarcity of comprehensive behavioral datasets for specific patient populations hampers the development of AI models capable of accurately detecting and monitoring health conditions through patient behavior analysis, especially in AD scenarios.

\vspace{-0.5em}
\subsection{Data Augmentation}

Data augmentation is widely used to enhance the diversity of training data~\cite{shorten2019survey}. Traditional data augmentation methods for activity recognition include spatial transformations (e.g., rotation, scaling), temporal manipulations (e.g., time warping), and adding noise to existing data~\cite{wang2018temporal}. While these techniques can introduce variability, they are constrained by the inherent characteristics of the original data and do not add new semantic information. In applications requiring the detection of subtle and complex behaviors, traditional data augmentation techniques may be insufficient. These methods cannot generate new behavior patterns or capture the nuanced differences needed for accurate recognition. \add{For instance, applying simple rotation or noise injection to a normal elder’s “standing up from a chair” action cannot realistically induce the repeated armrest supports or sustained stooping that real AD patients often exhibit (Figure~\ref{fig:example}). These nuanced shifts in posture or support patterns cannot be generated by basic transformations of healthy data.} Consequently, models trained with traditionally augmented data may fail to generalize well to real-world scenarios where such subtle behaviors are present~\cite{xu2022globem}.

\vspace{-0.5em}
\subsection{LLMs for Data Generation}

Recent advancements in generative models, particularly those leveraging LLMs and vision-language models, have shown promise in generating synthetic data~\cite{chen2024videocrafter2overcomingdatalimitations, cai2023generativerenderingcontrollable4dguided, guo2024vivargenerativearintuitive}. These models are trained on extensive datasets of common daily activities and general knowledge, enabling them to generate realistic content in these domains. \add{For example, RFGenesis~\cite{RFgenesis_ZXY} proposes a cross-modal diffusion framework to generate mmWave sensing data and improve zero-shot generalization, while UniMTS~\cite{zhang2024unimts} introduces a unified pre-training procedure for motion time series. These works highlight the growing trend of using generative models to enhance real-world data collection for Human Action Recognition (HAR).} 
However, in disease domains like AD, these generative models perform inadequately due to a lack of domain-specific knowledge~\cite{harmony}. The actions associated with AD patients are often subtle, fine-grained, and context-dependent, making them hard to capture without specialized understanding. 

Moreover, the inability to control the generated content at a fine-grained level poses challenges in tailoring the synthetic data to reflect the unique characteristics of AD-related actions. This gap underscores the need for generative models that can incorporate domain-specific knowledge and allow for precise control over the generated behaviors in smart health applications.

%% file: sections/system.tex
\section{System Design}\label{sec: system}

\begin{figure}[t]
\centering
\includegraphics[width=1.03\linewidth]{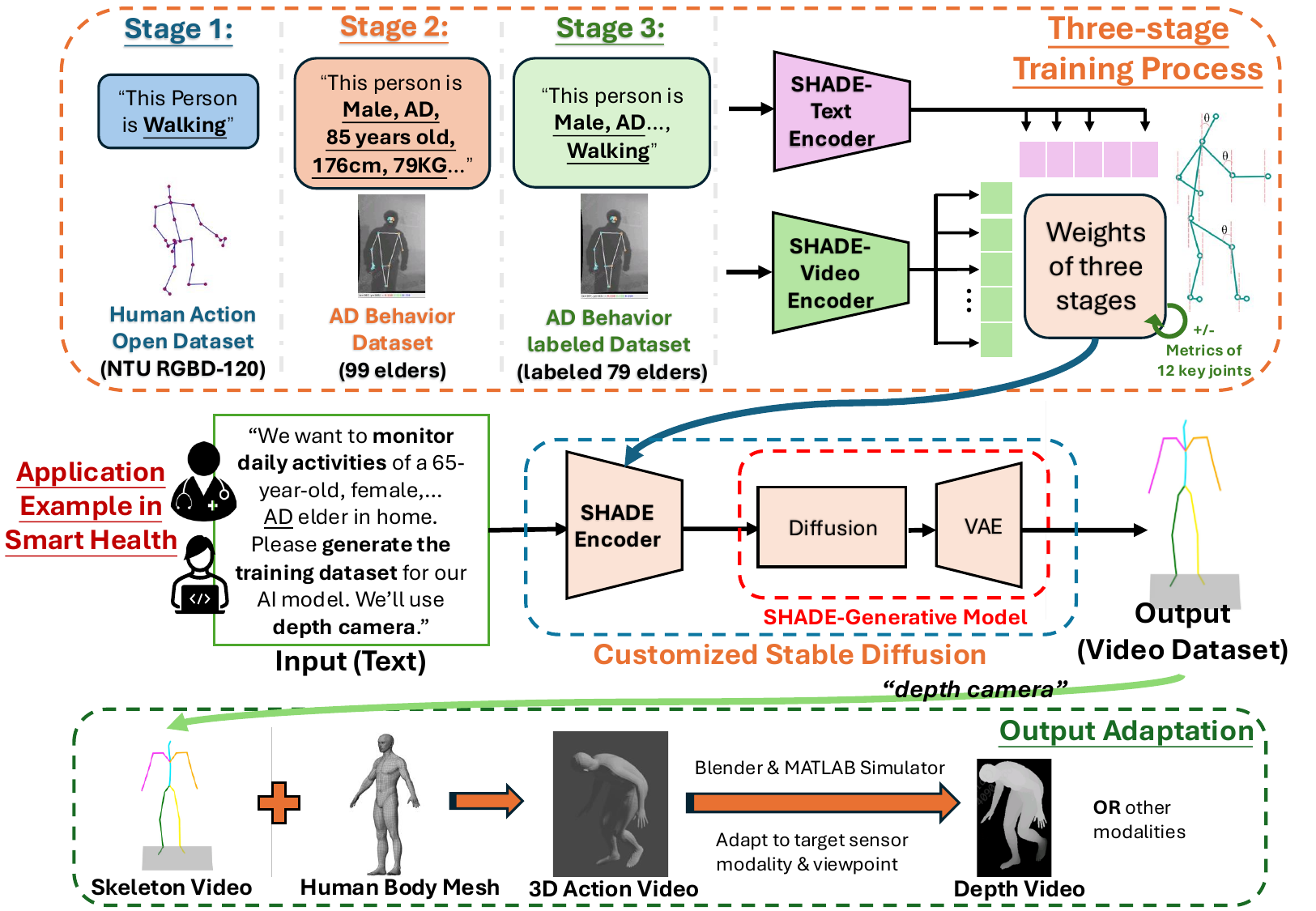}
\captionsetup{skip=0pt}
\caption{Design of SHADE-AD. The Training Process involves three stages: Stage 1 learns general human actions like \textit{"Walking"}; Stage 2 embeds AD-specific knowledge like \textit{"AD"} or \textit{"MoCa score"}; and Stage 3 fine-tunes with patient-specific motion metrics derived from 12 key joints. }
\label{fig:Pipeline}
\vspace{-2em}
\end{figure}

SHADE-AD integrates advanced components to model both general human activities and the subtle nuances of AD-specific behaviors. The framework and the operational workflow of SHADE-AD are illustrated in Figure~\ref{fig:Pipeline}, highlighting the interactions between its modules and the three-stage training procedure.

\subsection{System Components}

The SHADE-AD framework consists of three primary components: SHADE-Text Encoder, SHADE-Video Encoder and SHADE-Generative Model. Each component is meticulously designed to capture and integrate both general human actions and the subtle behavioral characteristic of AD.

\noindent
\textbf{SHADE-Text Encoder:} 
This module is responsible for processing textual descriptions and converting them into rich semantic embeddings. It utilizes a pre-trained OpenCLIP text encoder~\cite{radford2021learning}, but tailored for our application. To embed the encoder with sensitivity to AD-specific terminology and nuances, we fine-tune it on a specialized corpus containing medical assessments, cognitive descriptors (e.g., Montreal Cognitive Assessment (MoCa) score, Zarit Burden Interview (ZBI) score, etc.), and behavioral observations related to AD we collected from 99 patients. This fine-tuning enables the encoder to grasp the intricate language patterns and context associated with AD behaviors, ensuring that generated videos accurately reflect the specified conditions.

\noindent
\textbf{SHADE-Video Encoder:} 
The video encoder processes input activity videos to extract meaningful spatiotemporal representations. It combines CNNs for spatial feature extraction with temporal modeling~\cite{bertasius2021space} to capture motion dynamics over time. By training on both general action datasets and our AD-behavior dataset, the encoder learns to distinguish and represent the subtle differences in activity patterns associated with AD. The three-stage training approach enhances the encoder’s ability to capture a wide range of human actions while being sensitive to AD-specific characteristics.

\noindent
\textbf{SHADE-Generative Model:} Serving as the core of the SHADE-AD framework, the generative model synthesizes skeleton-based human action videos conditioned on textual prompts. We have developed a customized diffusion-based generative model~\cite{rombach2022high} that operates efficiently in a latent space, allowing for skeleton action generation. By integrating the embeddings from the text encoder and leveraging the spatiotemporal representations from the video encoder, the generative model produces videos that are both semantically aligned with the input text and exhibit activity patterns indicative of AD. 

\noindent
\textbf{Output Adaptation:} The generated skeleton video can be adapted to different sensor modalities using simulation tools like MATLAB and Blender, making it suitable for various smart health applications.

\subsection{Training Procedure}
\noindent \textit{Stage 1 - Pre-training on General Action Data}. 
In the first stage, we pre-train SHADE Encoder using a large-scale dataset of general human actions. Specifically, we utilize the NTU RGB+D 120 dataset~\cite{liu2019ntu}, which contains over 120,000 video samples across 120 different action classes. This dataset provides a comprehensive representation of common human activities, allowing our encoders to learn rich semantic and spatiotemporal features. \add{To preserve the broad linguistic knowledge within the OpenCLIP text encoder, most layers remain fixed. We enable gradient updates only on its final transformer block and projection layer to minimally adapt the encoder to action-specific terminology. Meanwhile, the video encoder is fully trained on open dataset, ensuring the two encoders remain well-aligned.}
SHADE-Text Encoder is trained to capture the semantic meanings of action descriptions, employing contrastive learning to align textual embeddings with corresponding video embeddings produced by SHADE-Video Encoder. 

\noindent \textit{Stage 2 - Fine-tuning on AD Behavior Dataset}.
We have collected a dataset involving 99 elders in various AD conditions, each monitored over a period of one month. Using privacy-preserving depth cameras installed in their primary living spaces, we obtained approximately 72,000 hours of video data, totaling around 100 TB.

\begin{figure}[h]
\vspace{-0.1em}
\centering
\includegraphics[width=1\linewidth]{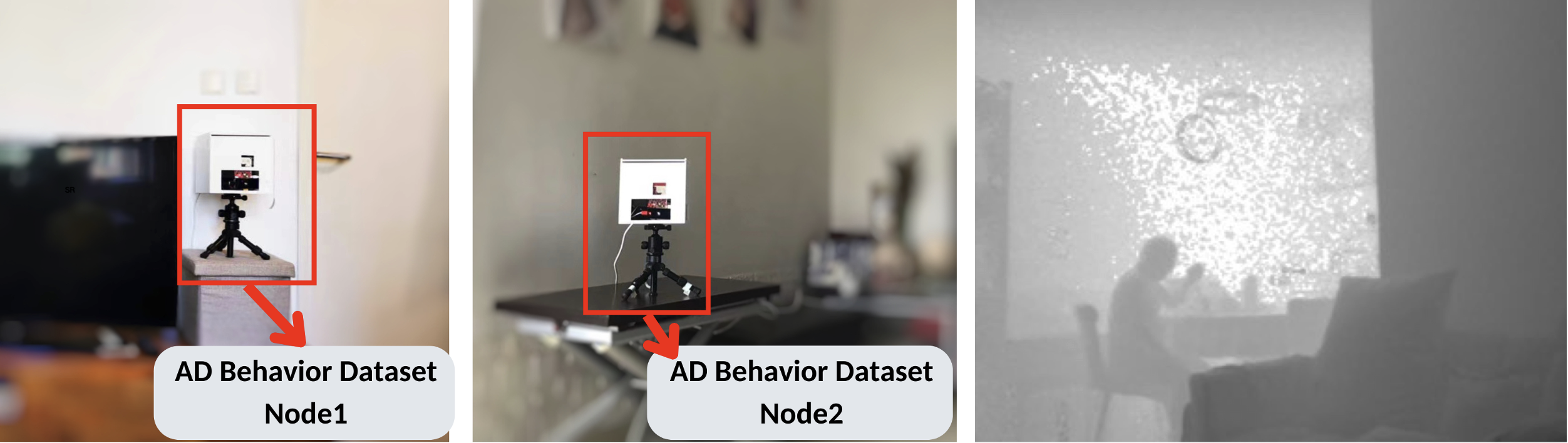}
\captionsetup{skip=0pt}
\caption{Examples of the layout of our data collection nodes and depth data in elders' homes.}
\label{fig:layout}
\vspace{-0.9em}
\end{figure}

This dataset is enriched with detailed annotations, including cognitive health statuses (e.g., AD, Mild Cognitive Impairment (MCI) and Normal Cognition (NC)) and medial assessments like MoCA scores. The annotations provide valuable context for understanding the severity and progression of AD-related behaviors. To process the raw video data into a format suitable for training and reduce computational overhead, we employ pose estimation algorithms~\cite{mmpose2020} to extract skeleton representations of the individuals.

During fine-tuning, we apply domain adaptation technologies~\cite{ganin2016domain, tzeng2017adversarial} to bridge the gap between the general action domain and the AD-specific domain. By introducing a domain discriminator and utilizing adversarial learning, the encoders learn domain-invariant features that are sensitive to AD-specific characteristics. The domain classification loss, denoted as \( \mathcal{L}_{\text{domain}} \), measures the accuracy of predicting the domain (general or AD-specific) of the extracted features.
\begin{equation*}
\mathcal{L}_{\text{domain}} = \text{CrossEntropy}(d_{\text{true}}, d_{\text{pred}})
\end{equation*}
\begin{equation*}
    \mathcal{L}_{\text{Stage2}} = \mathcal{L}_{\text{main}} - \lambda \cdot \mathcal{L}_{\text{domain}}
\end{equation*}

The total loss for the encoders, \( \mathcal{L}_{\text{total}} \), combines the main task loss (diffusion-based generation loss that ensures image quality and coherence) and the adversarial domain loss with a weighting factor \( \lambda \).
Here, the negative sign before \( \mathcal{L}_{\text{domain}} \) indicates the adversarial nature of the training, as the encoders aim to maximize this loss to make it harder for the domain discriminator to correctly identify the origin domain of the features, while the domain discriminator aims to minimize it by becoming more accurate at distinguishing between general and AD-specific domains. \add{Here, we adopt a smaller learning rate ($5 \times 10^{-5}$) to avoid catastrophic forgetting of general language features while embedding AD-specific concepts (e.g., ``MoCa Score= 15''). Concurrently, the video encoder undergoes adversarial alignment, ensuring it captures subtle AD cues, such as slower joint velocities. The domain alignment weight ($\lambda$) is set to~0.25.}

\noindent
\textit{Stage 3 - Fined-Grained Fine-Tuning with Patient-Specific Motion Metrics}.
We perform a fine-grained fine-tuning of the model using a subset of our AD behavior dataset that has been meticulously labeled. We selected data from 79 patients with rich action classes and labeled approximately 1\% of the data focusing on 22 AD-related activities.

We not only use conventional training losses but also incorporate motion metrics of 12 key joints as part of the loss function. These joints include the left and right hips, knees, ankles, elbows, shoulders, neck, and back. The goal is to ensure that the generated actions accurately reflect patient-specific physical characteristics. A detailed explanation of our motion metrics is given in the next section.

We introduce a motion metric loss $\mathcal{L}_{\text{metric}}$ that measures the discrepancy between the motion metrics of the generated data and those of the real patient data:
\begin{equation*}
\mathcal{L}_{\text{metric}} = \frac{1}{N} \sum_{i=1}^{N} \left\| M_{\text{real}}^{(i)} - M_{\text{synthetic}}^{(i)} \right\|_2^2
\end{equation*}
where $M_{\text{real}}^{(i)}$ and $M_{\text{synthetic}}^{(i)}$ represent the motion metrics of the $i$-th joint in the real and synthetic data, respectively, and $N$ is the number of key joints.
The overall loss function during Stage 3 combines the generative loss $\mathcal{L}_{\text{gen}}$ (e.g., adversarial loss, reconstruction loss) and the motion metric loss:
\begin{equation*}
\mathcal{L}_{\text{Stage3}} = \mathcal{L}_{\text{gen}} + \alpha \mathcal{L}_{\text{metric}}
\end{equation*}
where $\alpha$ is a weighting factor that balances the contribution of the motion metric loss. \add{We use a modest weighting factor (0.5) to balance AD realism and generative quality, preventing the model from overlooking subtle AD behaviors such as repeated support movements, while also achieving consistent performance as observed during training.}

By integrating the motion metrics into the training process, the model learns to generate synthetic data that not only appears realistic but also faithfully replicates the specific activity patterns and physical characteristics of AD patients.

Moreover, we implement a curriculum learning strategy~\cite{bengio2009curriculum} by gradually increasing the complexity of training samples. Starting with simpler actions commonly affected by AD (e.g., \textit{walking}, \textit{sitting}), we progressively introduce more complex behaviors. This strategy helps the model to stabilize during training and improves its capacity to capture subtle AD-related movement deviations.

\subsection{Advantages of the Design}

The SHADE-AD framework offers several advantages: 
\textit{1) AD-Specific Knowledge Embedding:} three-stage training mechanism embeds domain-specific knowledge, capturing AD patients’ subtle characteristics.
\textit{2) Adaptability to Sensor Modalities:} Synthetic skeleton videos can be adapted for various applications.
\textit{3) Privacy-Preserving and Cost-Effective:} Reduces the labeling work for real-world data.

\add{An example can be seen in Figure~\ref{fig:example}, where the our approach (prompted with ``an Alzheimer's Disease patient standing up from a chair'') demonstrates both the repeated support movements and the stooped posture, effectively capturing the motion characteristics observed in real AD patient data.} SHADE-AD framework effectively addresses the challenges of synthesizing AD-specific human activity videos, providing a valuable tool for enhancing smart health.

%% file: sections/validation.tex
 \section{Motion Metrics for Validation}\label{sec: validation}

\begin{figure}[htbp]
\centering
\includegraphics[width=0.95\linewidth]{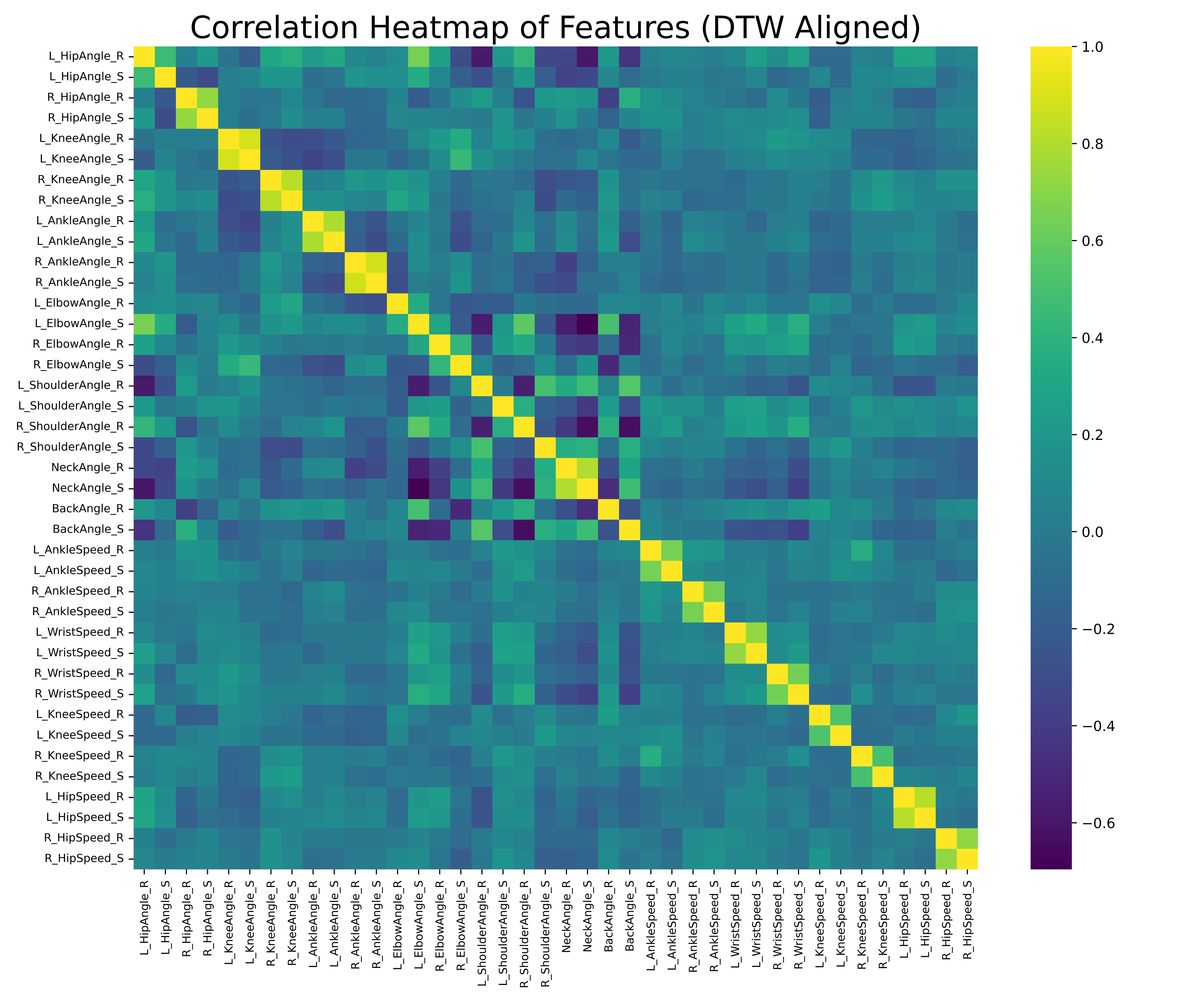} 
\captionsetup{skip=0pt}
\caption{Correlation heatmap of joint features between real and synthetic data (DTW aligned), demonstrating strong positive correlations for most features. Suffixes \emph{\_R} and \emph{\_S} represent Real and Synthetic data respectively.}
\vspace{-1em} 
\label{fig:CorrelationHeatmap}
\end{figure}

We performed a comprehensive comparative analysis focusing on 12 key joints of the human body to assess the similarities and differences between the real and synthetic data. The analysis employed various motion metrics, including joint speeds, joint angles, range of motion (ROM), and correlation measures, to evaluate how effectively the synthetic data replicates the motion patterns of real AD patients. We compare 200 \textit{"Walking"} action video clips between real elders' data and synthetic data generated using the elders' same health condition. Representive results are shown in Table~\ref{tab:motion_metrics_comparison} and illustrated by the correlation heatmap in Figure~\ref{fig:CorrelationHeatmap}.

\begin{table}[b]
\centering
\vspace{-1em}
\resizebox{1.0\columnwidth}{!}{%
    \begin{tabular}{lcccc}
    \toprule
    \textbf{Feature} & \textbf{Mean Ratio (S/R)} & \textbf{ROM Ratio (S/R)} & \textbf{Correlation} & \textbf{t-test p-value} \\
    \midrule
    \textbf{Left Hip Angle} & \textbf{0.72} & \textbf{1.24} & \textbf{0.8626} & $7.18 \times 10^{-3}$ \\
    \textbf{Right Hip Angle} & \textbf{0.84} & \textbf{0.88} & \textbf{0.7455} & \textbf{0.4487} \\
    \textbf{Left Knee Angle} & 0.68 & 1.70 & 0.5860 & $2.07 \times 10^{-4}$ \\
    \textbf{Right Knee Angle} & \textbf{0.81} & \textbf{0.84} & \textbf{0.7686} & \textbf{0.0655} \\
    \textbf{Left Shoulder Angle} & \textbf{1.17} & 0.49 & \textbf{0.8026} & 0.0009 \\
    \textbf{Right Shoulder Angle} & \textbf{0.91} & 0.60 & \textbf{0.7130} & \textbf{0.0502}\\
    \textbf{Neck Angle} & \textbf{0.99} & 0.63 & \textbf{0.8710} & \textbf{0.5028} \\
    Back Angle & \textbf{0.99} & \textbf{1.23} & 0.4323 & \textbf{0.3769} \\
    \midrule 
    Neck Speed & \textbf{0.95}& \textbf{0.84}& 0.6590 & 0.0029 \\
    \textbf{Left Ankle Speed} & \textbf{1.10} & \textbf{1.00} & \textbf{0.6444} & \textbf{0.6541} \\
    \textbf{Right Ankle Speed} & \textbf{0.83} & \textbf{0.83} & \textbf{0.6453} & \textbf{0.5045} \\
    \textbf{Left Wrist Speed} & \textbf{0.83} & \textbf{0.87} &\textbf{ 0.7241} & \textbf{0.7618} \\
\textbf{Right Wrist Speed}& \textbf{0.99} & \textbf{0.76} & \textbf{0.6389} & \textbf{0.5651} \\
    \textbf{Left Knee Speed} & \textbf{1.18} & \textbf{1.01} & \textbf{0.5260} & \textbf{0.4507} \\
    \textbf{Right Knee Speed} & \textbf{0.73} & \textbf{1.00} & \textbf{0.5000} & \textbf{0.3064} \\
    \textbf{Left Hip Speed} & \textbf{1.12} & \textbf{1.52} & \textbf{0.8167} & \textbf{0.6335} \\
    \textbf{Right Hip Speed} & \textbf{1.00} &\textbf{ 0.98} & \textbf{0.7214} & \textbf{0.9538} \\
    \bottomrule
    \end{tabular}
}
\caption{Comparison of Motion Metrics Between Real and SHADE-AD Synthetic Data on \textit{"Walking"} action.}
\vspace{-1.8em}
\label{tab:motion_metrics_comparison}
\end{table}

\begin{figure}[htbp]
\centering
\begin{subfigure}[t]{0.48\linewidth} 
    \centering
    \includegraphics[width=\linewidth]{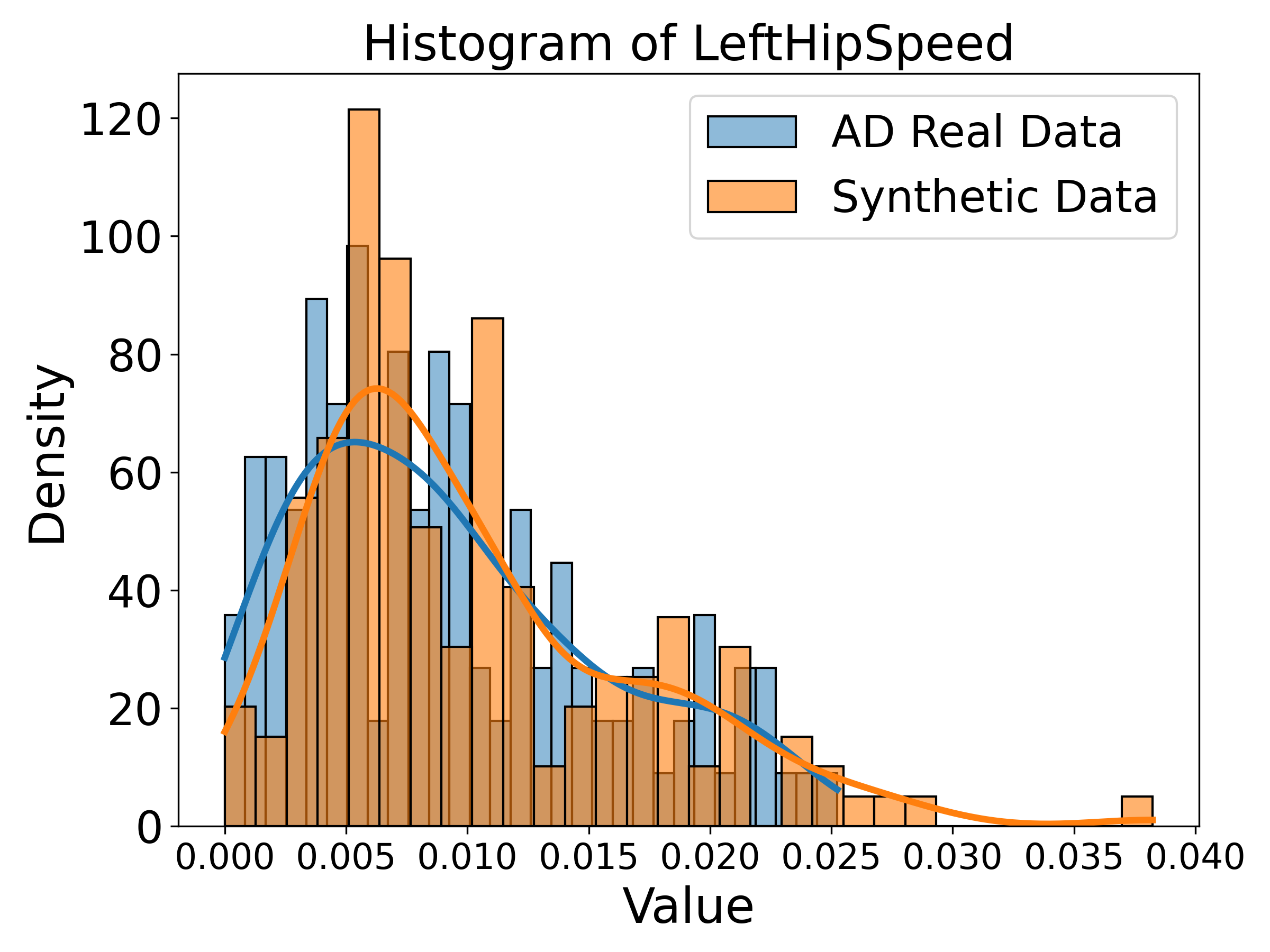}
    \label{fig:LeftHipSpeedHistogram}
\end{subfigure}
\hfill
\begin{subfigure}[t]{0.48\linewidth} 
    \centering
    \includegraphics[width=\linewidth]{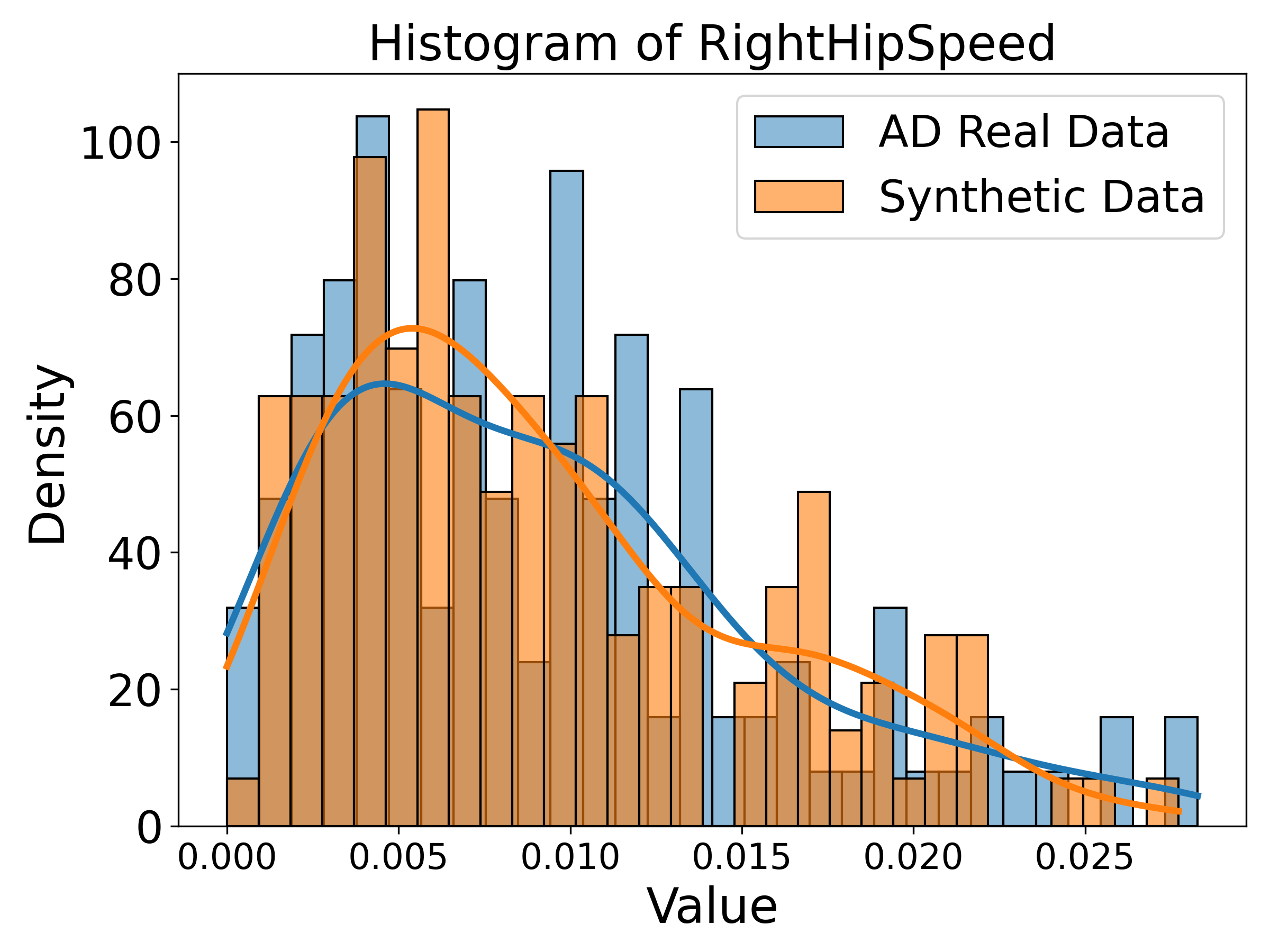}
    \label{fig:RightHipSpeedHistogram}
\end{subfigure}


\begin{subfigure}[t]{0.48\linewidth} 
    \centering
    \includegraphics[width=\linewidth]{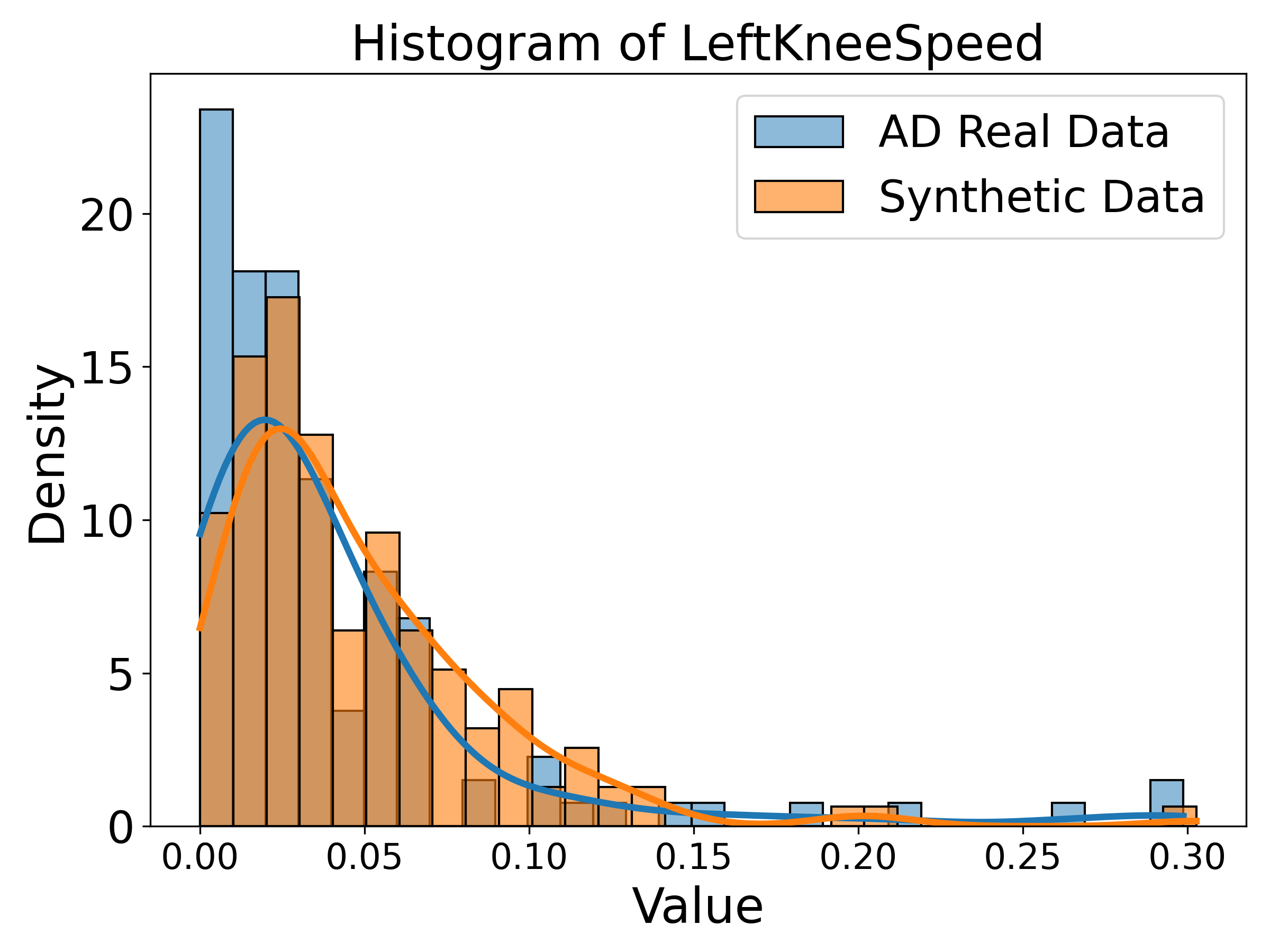}
    \captionsetup{skip=0pt}
    \label{fig:LeftKneeSpeedHistogram}
\end{subfigure}
\hfill
\begin{subfigure}[t]{0.48\linewidth} 
    \centering
    \includegraphics[width=\linewidth]{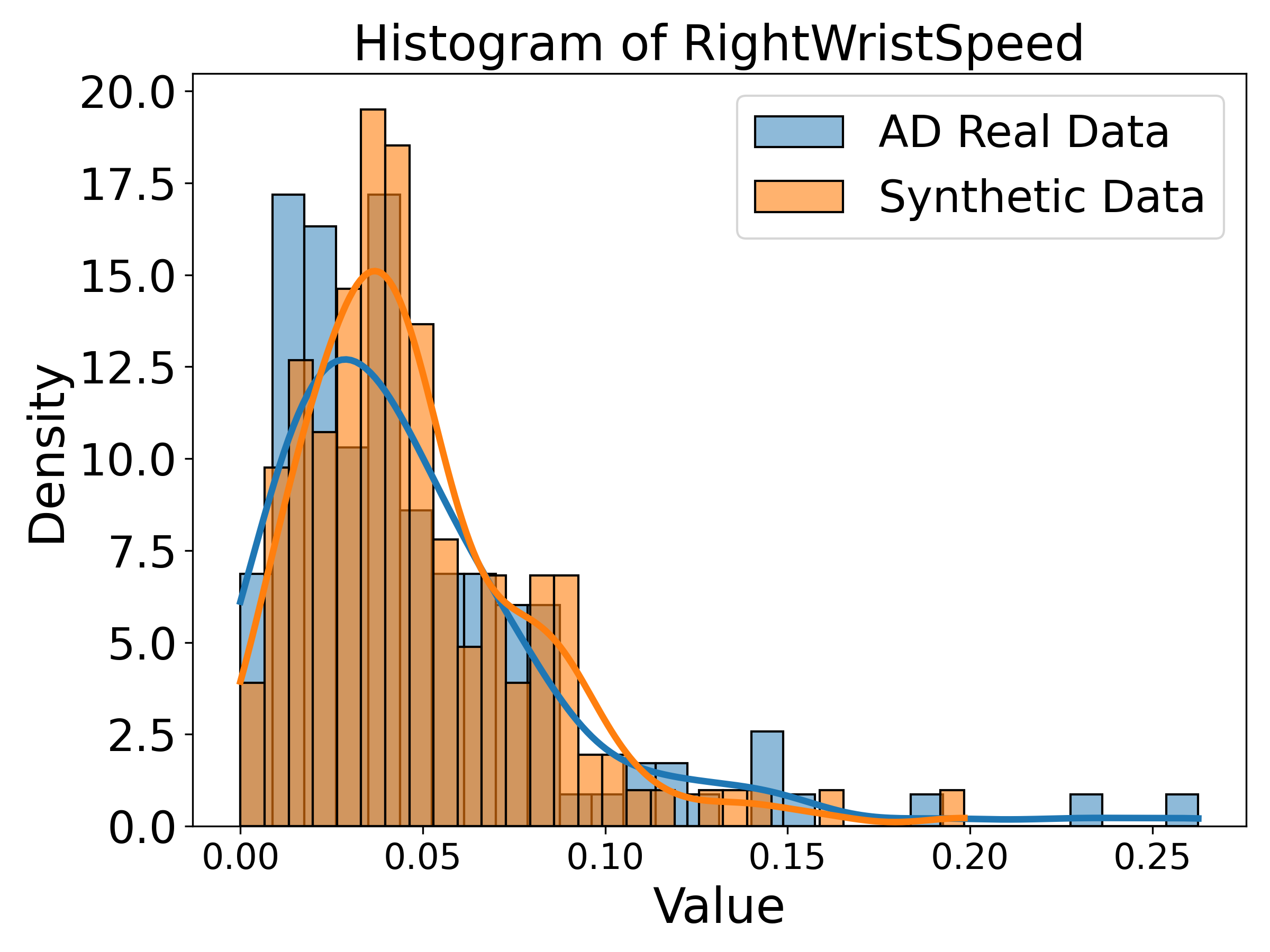}
    \captionsetup{skip=0pt}
    \label{fig:RightWristSpeedHistogram}
\end{subfigure}
\vspace{-0.7em}
\captionsetup{skip=0pt}
\caption{Histograms of various features for real and synthetic data, showing overlapping distributions and indicating similar movement dynamics.}
\vspace{-1em} 
\label{fig:FourHistograms}
\end{figure}

\begin{figure}[h]
\centering
\begin{subfigure}[t]{0.47\linewidth}
    \centering
    \includegraphics[width=\linewidth]{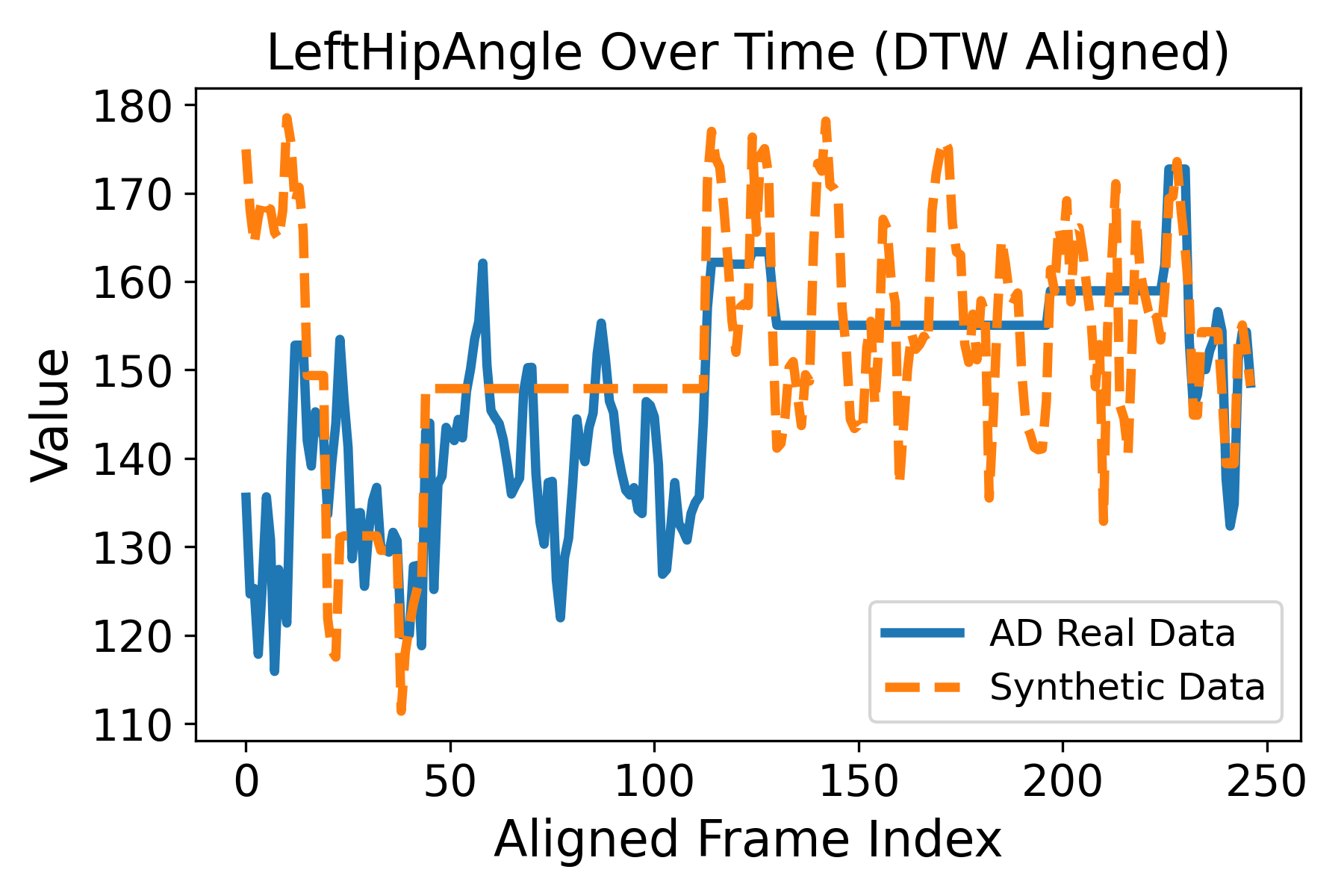}
    \label{fig:LeftHipAngleTimeSeries}
\end{subfigure}%
\hspace{0.05\linewidth} 
\begin{subfigure}[t]{0.47\linewidth}
    \centering
    \includegraphics[width=\linewidth]{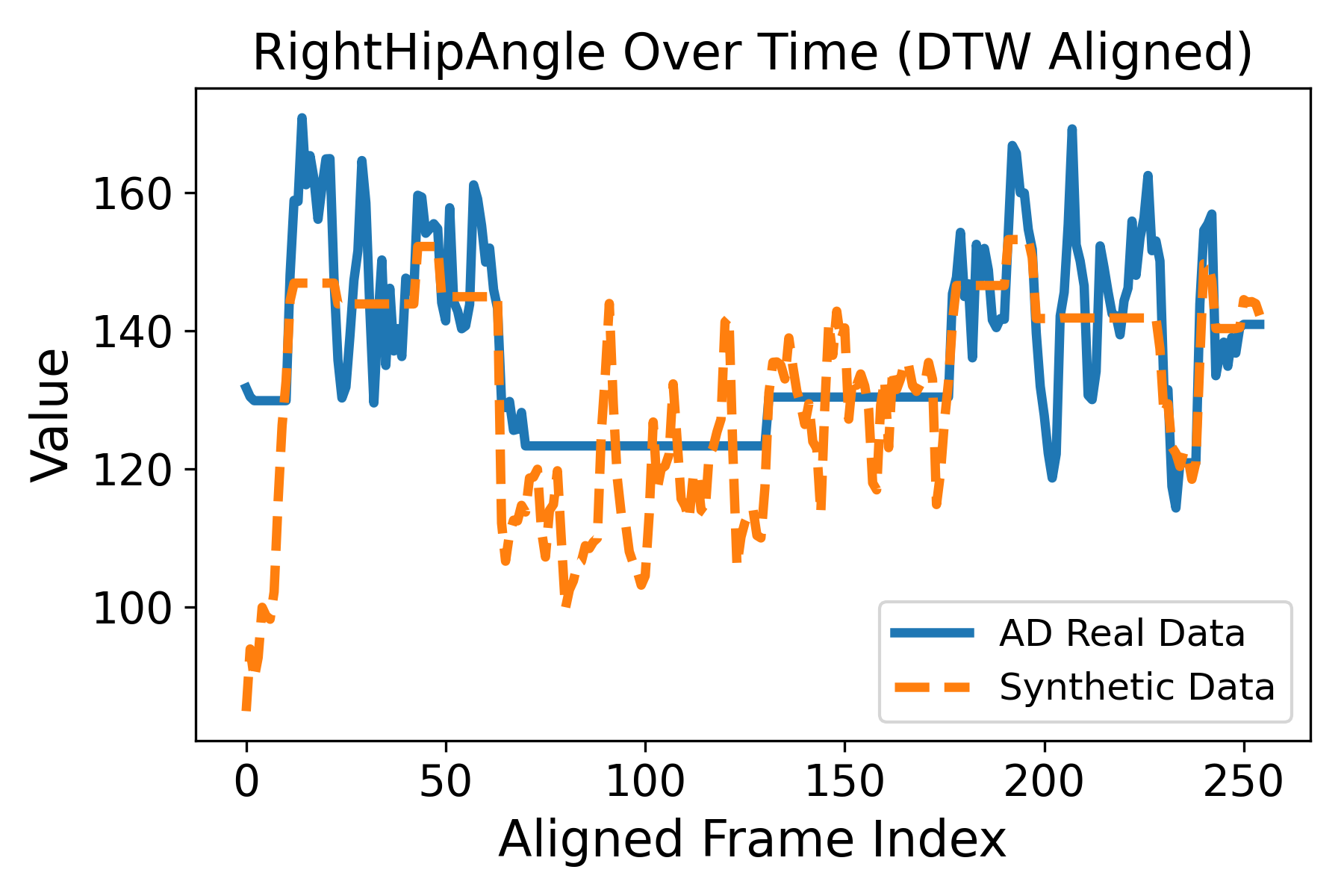}
    \label{fig:RightHipAngleTimeSeries}
\end{subfigure}


\begin{subfigure}[t]{0.47\linewidth}
    \centering
    \includegraphics[width=\linewidth]{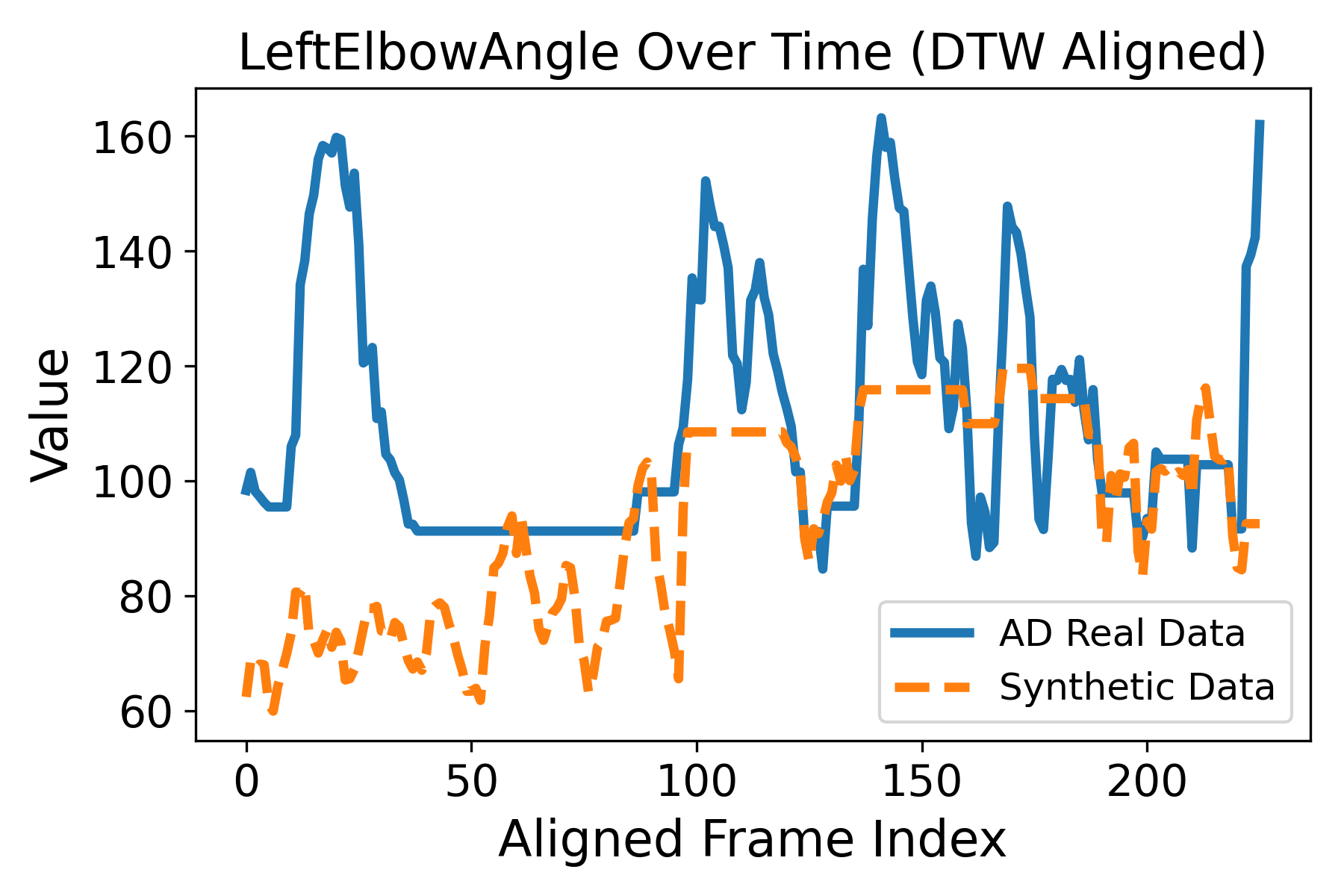}
    \label{fig:LeftElbowAngleTimeSeries}
\end{subfigure}%
\hspace{0.05\linewidth} 
\begin{subfigure}[t]{0.47\linewidth}
    \centering
    \includegraphics[width=\linewidth]{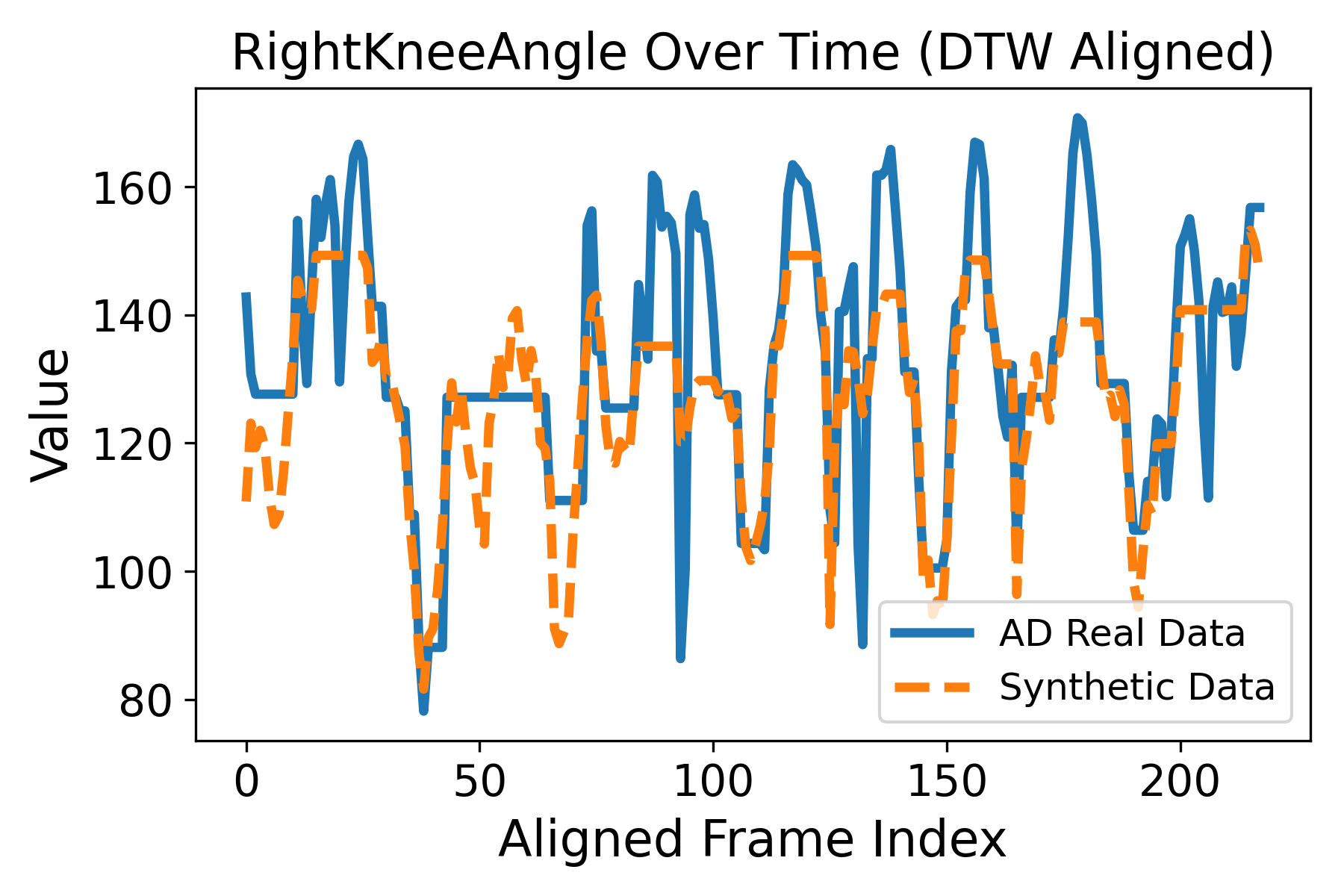}
    \label{fig:RightKneeAngleTimeSeries}
\end{subfigure}

\captionsetup{skip=0pt}
\caption{Time series of joint speeds and angles over time (DTW aligned) for real and synthetic data, demonstrating consistent trends within action cycles.}
\vspace{-1.5em}
\label{fig:JointTimeSeriesDTW}
\end{figure}

\noindent \textit{Joint Speed Distribution.} The distributions of joint speeds between the real and synthetic data show a high degree of overlap across the analyzed joints. For instance, the histograms of \textit{Left Hip Speed} and \textit{Right Wrist Speed} (Figures~\ref{fig:FourHistograms}) illustrate that the synthetic data effectively replicates the speed characteristics of AD patients. This close match suggests that SHADE-AD successfully learned the slower movement speeds typical of AD-related motor decline. This is critical for recognizing actions like \textit{walking}, where speed variations are significant indicators of motor impairment.

\noindent \textit{Joint Angle Trajectories.} The time series of joint angles, after DTW alignment, exhibit similar patterns and trends between the real and synthetic data in one activity period. As shown in Figure~\ref{fig:JointTimeSeriesDTW}, the \textit{Right Knee Angle} over time demonstrates consistent cyclical movements corresponding to gait patterns. Although there are minor discrepancies at the initiation of actions, likely due to differences in starting postures, the overall alignment indicates that the synthetic data captures the essential movement dynamics.

\noindent \textit{Mean and Range of Motion (ROM).} Most Mean and ROM Ratios of synthetic and real data are around 1.0 in Table~\ref{tab:motion_metrics_comparison}, which confirms that the synthetic data accurately represents the limitations in joint mobility associated with AD.

\noindent \textit{Correlation.} The correlation coefficients between real and synthetic data are generally high for most features, with several exceeding 0.7, indicating strong positive relationships (Table~\ref{tab:motion_metrics_comparison}). The correlation heatmap (Figure~\ref{fig:CorrelationHeatmap}) visually reinforces this finding, showing strong correlations across multiple joint features.

\noindent \textit{Statistical Tests.} The t-test p-values provide statistical evidence regarding the differences between the datasets. For features like \textit{Right Hip Angle}, high p-values (p > 0.05) indicate no significant difference between the real and synthetic data. This statistical similarity further supports the conclusion that the gap between the datasets is minimal.

%% file: sections/result.tex
\section{Preliminary Results}\label{sec:result}

To assess the effectiveness of the proposed SHADE-AD framework, we first compare our method with four existing data enhancement approaches, and subsequently evaluate the adaptability of SHADE-AD across another data modality.

\subsection{Experiment 1: Human Activity Recognition Task (Skeleton-based)}\label{exp1}

\noindent\textbf{Objective:} Assess the performance of HAR models trained with our synthetic data and other data enhancement methods in recognizing activities in AD patients.

\noindent\textbf{Model and Methods:} We trained a lightweight HAR model composed of 5 CNN layers and 2 RNN layers, suitable for deployment on embedded systems in smart health applications. The model was trained with following methods:

\noindent\textit{1) Vanilla Model without data augmentation (Model V)}. Trained on AD behavior dataset containing 21,910 labeled skeleton videos from 79 elderly individuals, with a biased distribution heavily favoring the \textit{sitting} action (\textit{sitting}: 16,596; \textit{standing}: 3,121; \textit{walking}: 1,735; \textit{turning}: 272; \textit{lying}: 186).

\noindent\textit{2) Model enhanced with common Data Augmentation (Model DA)}. Trained on the AD behavior dataset using common data augmentation techniques (rotation, scaling and noise addition), increasing the dataset size by 1.5 times.

\noindent\add{\textit{3) Model enhanced with common Statistical Method (Model SM)}. Trained on the AD behavior dataset using bootstrapping strategy that resamples skeleton sequences, increasing the dataset size by 1.5 times.}

\noindent\textit{4) Model enhanced with Open Dataset (Model OD)}. Trained on the AD behavior dataset combined with data from open action dataset. Incorporated 5,000 skeleton videos from the NTU RGB+D 120 dataset~\cite{liu2019ntu}, covering similar action classes performed by healthy individuals.

\noindent\textit{5) Model enhanced with LLM-Generated Data (Model LLM)}. Trained using synthetic data generated by MotionDiffuse~\cite{zhang2022motiondiffuse}, containing 20,000 videos with balanced action distribution.

\noindent\add{\textit{6) Model enhanced with LLM and Text Conditioning (Model LLM-AD)}. Similar to 5) but with prompts explicitly mentioning “AD elder”.} 

\noindent\textit{7) Model enhanced with Our Synthetic Data (Model OS)}. Trained using synthetic data generated by SHADE-AD, which contains 20,000 videos with
balanced action distribution, evenly health condition representation (AD, MCI, and NC), and 360-degree viewpoints around subjects. Prompts are structured like, “Generate a HAR training dataset for an elder with AD, MoCa Score= 9, ..., of walking action, from a front view angle.”

\noindent\textbf{Testing Data:} We used 3,200 action videos from another 20 elders. To test the real-world performances of the five approaches more precisely, we select this cross-subject dataset with a balanced action distribution to mitigate the problems of high variance and high bias.


\begin{table}[b]
    \centering
    \vspace{-1.5em}
    \resizebox{1.0\columnwidth}{!}{%
    \begin{tabular}{lccccc}
    \toprule
    \textbf{Model} & \textbf{Sitting (\%)} & \textbf{Standing (\%)} & \textbf{Walking (\%)} & \textbf{Turning (\%)} & \textbf{Lying (\%)} \\
    \midrule
    Model V (base) & 100\% & 0\% & 3.13\% & 0\% & 0\% \\
    Model DA & 96.25\%& 10.63\% & 8.75\% & 2.03\% & 1.56\% \\
    \add{Model SM} & 97.34\%& 14.22\%& 10.63\%& 4.84\%& 2.50\%\\
    Model OD & 83.28\% & 25.16\% & 12.19\% & 15.16\% & 8.44\% \\
    Model LLM & 69.06\%& 66.25\%& 64.38\% & 63.28\% & 58.75\% \\
    \add{Model LLM-AD} & 72.34\%& 67.97\%& 68.91\%& 70.16\%& 60.47\%\\
    \textbf{Model OS} & \textbf{76.56\%} & \textbf{73.44\%}& \textbf{78.13\%}& \textbf{79.69\%}& \textbf{75.00\%}\\
    \bottomrule
    \end{tabular}
    }
    \caption{HAR Accuracies using different data enhancement methods on testing data.}
    \label{tab:har_comparison}
    \vspace{-1.5em}
\end{table}

\noindent\textbf{Analysis:} The results of Experiment 1 shown in Table~\ref{tab:har_comparison} highlight the limitations of existing methods and the effectiveness of our SHADE-AD approach:
\textit{Model V} achieves 100.00\% accuracy for the \textit{sitting} action due to severe data imbalance but performs poorly on other actions, indicating overfitting and inability to generalize; 
\textit{Model DA} slightly improves recognition of underrepresented actions, but the enhancements are minimal because these techniques do not introduce new semantic information or address the complexity of AD-specific behaviors; 
\add{\textit{Model SM} performs marginally better than \textit{Model DA}. By re-sampling skeleton sequences, it diversifies the training data, yet still fails to generate fundamentally new disease-relevant behavior patterns (e.g., subtle stooping);}
\textit{Model OD} yields better performance on some common actions; however, discrepancies in action execution and context—such as different gait patterns and movement nuances between healthy individuals and AD patients—result in poor recognition; 
\textit{Model LLM} shows moderate improvements across actions, yet it lacks AD-specific features since current generative models do not capture the subtle, disease-related action patterns; 
\add{\textit{Model LLM-AD} improves upon \textit{Model LLM} by including “AD elder” in the prompts, leading to better recognition on walking and turning—two actions that can manifest subtle AD characteristics;}
\textit{Model OS}, trained with synthetic data generated by SHADE-AD that embeds AD-specific behavioral knowledge, achieves balanced and significantly high accuracies across all action classes. This result demonstrates that our approach effectively addresses data imbalance, captures complex AD-related behaviors, and enhances model generalization by providing realistic and diverse training examples tailored to the specific characteristics of AD patients.

\subsection{Experiment 2: Test the Adaptability to Different Modality (Depth Videos)}\label{exp4}

\noindent\textbf{Objective:} Evaluate the adaptability of our approach to another data modalities beyond skeleton data.

\noindent\textbf{Model and Methods:} A Depth-based HAR model is used.

\noindent \textit{1) Vanilla Depth Model without data augmentation (Model V-Depth)}. Trained on AD behavior depth videos (corresponding to the skeleton data in AD behavior dataset used in Exp. 1).

\noindent \textit{2) Our Enhanced Depth Model (Model OS-Depth)}. Trained on synthetic depth videos simulated from our synthetic skeleton data in Experiment. 1 using Matlab and Blender.

\vspace{-0.5em}
\begin{table}[h]
    \centering
    \resizebox{1.0\columnwidth}{!}{%
    \begin{tabular}{lccccc}
        \toprule
        \textbf{Model} & \textbf{Sitting (\%)} & \textbf{Standing (\%)} & \textbf{Walking (\%)} & \textbf{Turning (\%)} & \textbf{Lying (\%)} \\
        \midrule
        Model V-Depth & 97.97\% & 5.0\% & 0\% & 2.03\% & 0\% \\
        \textbf{Model OS-Depth} & \textbf{73.75\%} & \textbf{70.94\%} & \textbf{66.56\%} & \textbf{67.34\%} & \textbf{65.63\%} \\
        \bottomrule
    \end{tabular}
    }
    \caption{Depth-Based HAR Accuracies on the testing data of 3,200 raw depth videos same in Experiment 1.}
    \label{tab:depth_results}
    \vspace{-1.5em}
    \end{table}
\vspace{-1em}
\noindent\textbf{Analysis:} \textit{Model OS-Depth} significantly improves recognition, which demonstrates that SHADE-AD can be adapted to other sensor modalities, starting from skeleton data and processing it to match various data types required by edge devices. The ability to generate synthetic data suitable for different modalities highlights the flexibility and practicality of SHADE-AD in real-world applications.

\subsection{Highlights}

The enhanced performance of our models can be attributed to: \textit{1) Class Balance:} The synthetic dataset provided equal representation of all action classes, mitigating the bias present in the real-world data. \textit{2) Physiological Variation:} Inclusion of diverse physiological characteristics (e.g., AD, MCI and NC) in the synthetic data allowed the models to learn variations in action patterns associated with different health conditions. \textit{3) Viewpoint Diversity:} Synthetic data included multiple viewpoints, improving the models’ ability to recognize activities from different angles.

\subsection{Future Work}
\add{
While our approach enhances AD-specific HAR, further improvements can be explored. \textit{1) Performance Improvement:} Incorporating biomechanical constraints and broader datasets may enhance movement fidelity and generalization. \textit{2) Comprehensive Motion Validation:} Beyond key-joint metrics, multi-joint coordination analysis can provide a more comprehensive assessment. \textit{3) Semantic Consistency and AD Relevance:} Future refinements will focus on ensuring semantic consistency and clinical relevance in realistic disease progression.
}

%% file: sections/conclusion.tex
\section{CONCLUSION} \label{sec:conclusion}

We present SHADE-AD as a privacy-preserving and low-cost tool for synthesizing human activity datasets for AD patients.
SHADE-AD demonstrates a significant improvement in the accuracy of human activity recognition when enhancing embedded models using a synthesis dataset, which shows a promising potential in smart health.

%% file: sections/acknowledgment.tex
\section{Acknowledgment}\label{sec: acknowledgment}

This paper is supported by Hong Kong RGC GRF 14201924 and CUHK Direct Grant 4055216.